\documentclass[lettersize,journal]{IEEEtran}
\usepackage{amsmath,amsfonts}
\usepackage{algorithmic}
\usepackage{array}
\usepackage[caption=false,font=normalsize,labelfont=sf,textfont=sf]{subfig}
\usepackage{textcomp}
\usepackage{stfloats}
\usepackage{url}
\usepackage{verbatim}
\usepackage{graphicx}
\usepackage{multirow}
\usepackage{amssymb}
\usepackage{amsthm}
\usepackage{mathrsfs}

\usepackage{xcolor}
\usepackage{manyfoot}
\usepackage{booktabs}
\usepackage{listings}
\usepackage{cite}
\usepackage{balance}
\usepackage{tabularx}
\usepackage{makecell}
\usepackage{placeins}
\def\BibTeX{{\rm B\kern-.05em{\sc i\kern-.025em b}\kern-.08em
		T\kern-.1667em\lower.7ex\hbox{E}\kern-.125emX}}

\theoremstyle{plain}

\begin{document}
	
	\title{Towards Efficient Multimodal and Multilingual Opinion Extraction for STI: A QLoRA-Based Fine-Tuning Approach}
	
	\author{Sheng Hong, Xuanqi Wang, Jiacheng Wang, and Yuwei Wang%
		\thanks{Sheng Hong is with the School of Cyber Science and Technology, Beihang University, Beijing, 100191, China (e-mail: shenghong@buaa.edu.cn).}%
		\thanks{Xuanqi Wang is with the School of Information and Engineering, Nanchang University, Nanchang, 330031, China (e-mail: 15894886025@163.com).}%
		\thanks{Jiacheng Wang is with the School of Cyber Science and Technology, Beihang University, Beijing, 100191, China (e-mail: wjc1321@163.com).}%
		\thanks{Yuwei Wang is with the Institute of Computing Technology, Chinese Academy of Sciences, Beijing, 100190, China (e-mail: ywwang@ict.ac.cn).}%
		\thanks{These authors contributed equally to this work.}%
		\thanks{This work is supported by National Key Research and Development Program [2022YFB3103602] and the Ministry of Industry and Information Technology High Quality Project [ZC26T320064-111].}%
	}
	
	\markboth{IEEE Transactions on Big Data,~Vol.~XX, No.~XX, ~2024}%
	{Hong et al.: Towards Efficient Multimodal and Multilingual Opinion Extraction for STI}
	
	\maketitle
	
	\begin{abstract}
		Recent advances in large language models (LLMs) have reshaped semantic analysis. Opinion Extraction (OE) for Science and Technology Intelligence (STI) requires concise core opinions from large information streams. Off-the-shelf models struggle to filter noise from these streams and show limited structured-output reliability in zero-shot multilingual and multimodal settings. To address information overload and extraction defocus, this study proposes a multimodal core-opinion extraction framework in which visual evidence serves as a contextual anchor for textual judgment. Using VideoLLaMA2 (VL2) and VideoLLaMA2.1 (VL2.1) as the base models, we apply Quantized Low-Rank Adaptation (QLoRA) fine-tuning on a curated dataset of 2,194 multilingual and multimodal samples. Under the selected Image-Augmented setting, fine-tuned VL2.1 generates structured JSON core-opinion outputs, achieving 64.98\% Precision, 42.15\% Recall, 51.14\% F1-score, and 74.00\% sample-level accuracy. Relative to the zero-shot VL2.1 setting, it raises the F1-scores of Spanish and Russian from 4.83\% and 0.45\% to 46.05\% and 51.93\%, respectively. The framework further incorporates a Fuzzy Cumulative Prospect Theory-based post-extraction triage module for case-level value assessment, providing a case-level value signal for downstream STI screening.
	\end{abstract}
	
	
	\section{Introduction}
	\IEEEPARstart{S}{cience} and Technology Intelligence (STI) analysis aims to track technological innovations and public evaluations from large and heterogeneous information sources. As information dissemination becomes increasingly global, STI data also becomes more multimodal and multilingual. Major technological breakthroughs are often accompanied by news reports, social media comments, and video demonstrations in multiple languages, including Chinese, English, Russian, and Spanish.

	Recent work has begun to adapt large models to domain-specific evaluation and industrial analysis settings \cite{1,5,26}. Recent multimodal video analysis has also shown that fusing spatial, frequency-domain, and motion cues can improve robustness in complex media understanding tasks, which further supports the use of heterogeneous evidence in STI-oriented settings \cite{hong2026aigcvideo}. Related work on video object segmentation likewise indicates that sparse propagation and frame-relation mining can strengthen spatio-temporal robustness in dynamic visual scenes \cite{dang2023efficient}. In STI opinion extraction, multimodal large language models (MLLMs) face three challenges: extraction defocus, language drift, and resource constraints. Extraction defocus can yield broad sets of peripheral opinions and reduce the clarity of structured core-opinion outputs \cite{8,9}. Language drift can affect output stability in non-English settings such as Russian and Spanish \cite{7}. Resource constraints motivate the use of parameter-efficient fine-tuning (PEFT) strategies \cite{10,16}.
	
	To turn large language models (LLMs) from broad semantic readers into focused intelligence extractors, we propose a fine-tuning and evaluation framework built on VideoLLaMA2 (VL2) and VideoLLaMA2.1 (VL2.1). The framework treats multimodal information, including images and videos, as contextual anchors that help the model distinguish core opinions from peripheral statements in text. Extracting the core opinion, however, is only the first step; the resulting structured cases must also be assessed for downstream STI screening.
	
	This work makes three contributions. First, we construct a multimodal STI opinion dataset with 2,194 annotated instances across four languages, providing the visual and textual grounding required to train the model to focus on core opinions. Second, we apply Quantized Low-Rank Adaptation (QLoRA) to adapt a 7B-parameter model with parameter-efficient updates and constrain semantic analysis to JSON-formatted core-opinion outputs at the extraction stage; experiments show that fine-tuned VL2.1 in the selected Image-Augmented setting achieves 64.98\% Precision, 42.15\% Recall, 51.14\% F1-score, and 74.00\% sample-level accuracy, improving performance in lower-resource languages. Third, we introduce a Fuzzy-CPT-based case-level value assessment module that evaluates each extracted case across five dimensions and supports STI monitoring.

	\begin{figure*}[!t]
		\centering
		\includegraphics[width=0.96\textwidth]{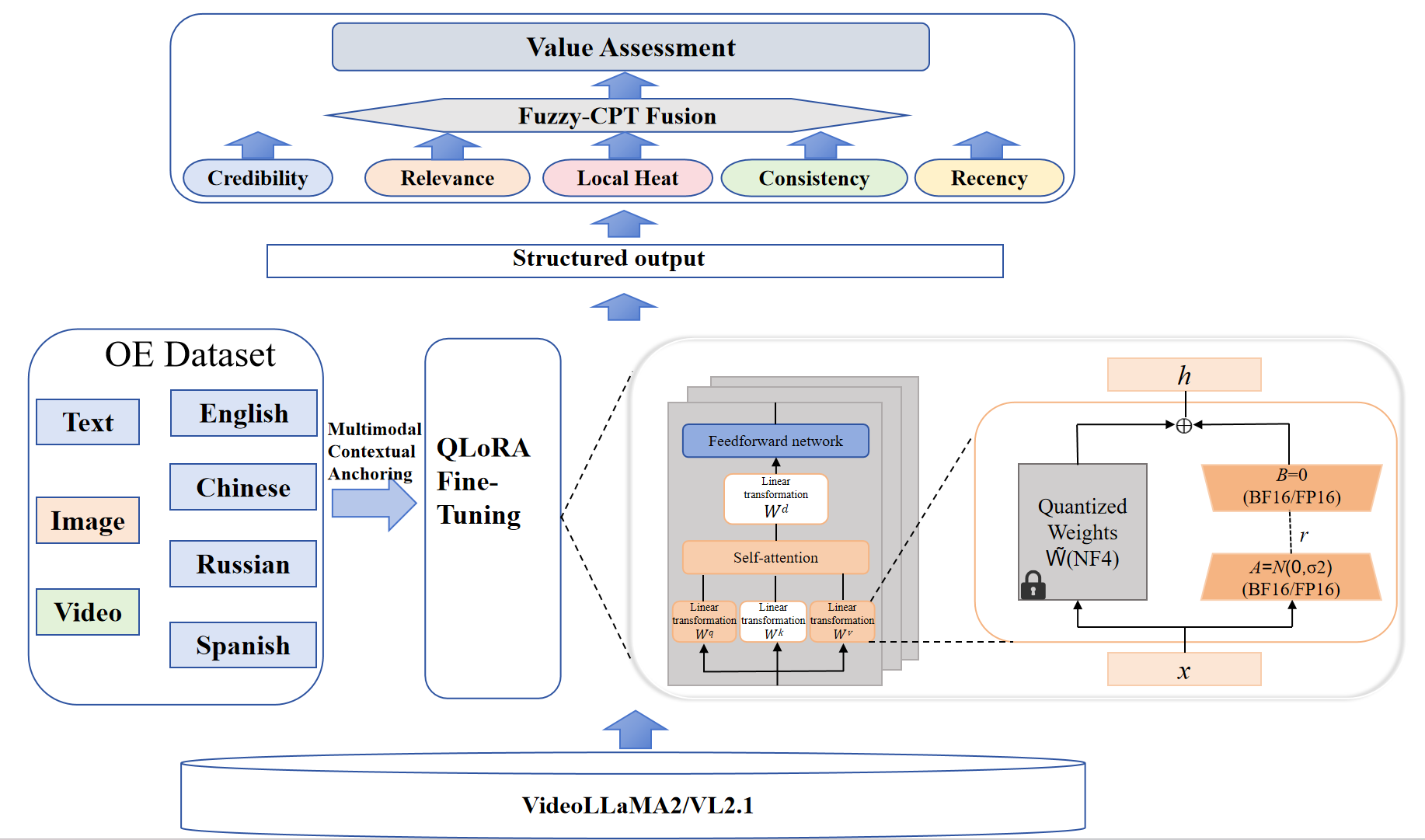}
		\caption{Overall framework of the proposed method.}
		\label{fig:framework}
	\end{figure*}
	
	\section{Methodology}
	\subsection{Overall Framework}
To bridge the gap between broad semantic reading and focused intelligence extraction, we design an integrated framework for multimodal and multilingual core-opinion extraction in STI. The framework transforms unstructured and noisy data streams into structured core opinions and screening-oriented intelligence cues. As illustrated in Fig.~\ref{fig:framework}, the pipeline consists of four components.
	
The pipeline begins with multimodal data anchoring through a self-constructed STI dataset with 2,194 annotated samples across four languages (English, Chinese, Spanish, and Russian) and three modalities (text, image, and video). By aligning visual cues with textual statements, this multimodal foundation helps the model ground subjective opinions and reduce drift toward irrelevant details. Visual cues provide contextual evidence that helps the model identify the core opinion expressed in text. This design keeps temporal evidence lightweight throughout the pipeline \cite{dang2025fwise}. QLoRA fine-tuning then serves as the main focusing mechanism. By adapting the 7B-parameter VL2/VL2.1 backbones in 4-bit form, this stage encourages the model to suppress peripheral noise and retain core opinions relevant to STI analysis. The resulting model maps the focused semantic representation into formatted JSON core opinions, reducing the structural inconsistency that is common in generative outputs and facilitating integration with STI monitoring databases. Finally, the case construction process organizes each predicted opinion with its source, temporal information, and subject context. The Fuzzy-CPT module then evaluates the case across credibility, relevance, local heat, consistency, and recency.
	
	\subsection{Dataset Construction}
\begin{figure}[!t]
		\centering
		\includegraphics[width=0.5\columnwidth]{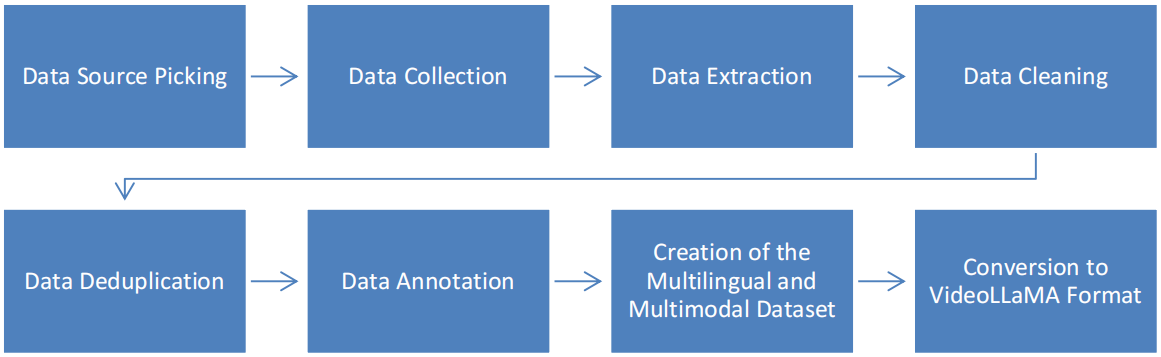}
		\caption{Construction process of the proposed dataset.}
		\label{fig:dataset_construction}
	\end{figure}

	We construct a new multimodal and multilingual dataset for core-opinion extraction in STI. Available multilingual opinion-related datasets are typically language-specific or task-specific \cite{18,19,31}. Although the raw collection pipeline may encounter text, images, videos, and occasional audio-bearing pages, the finalized OE dataset used in this study retains only three modalities: text, image, and video. The dataset spans four languages (English, Chinese, Spanish, and Russian), selected according to geopolitical relevance and data availability. To support robust STI core-opinion extraction experiments on subjective content, the dataset is built through a structured eight-step process: (1) Data Source Picking, (2) Data Collection, (3) Data Extraction, (4) Data Cleaning, (5) Data Deduplication, (6) Data Annotation, (7) Creation of a Multilingual and Multimodal Dataset, and (8) Conversion to VL2 Format. The construction process is illustrated in Fig.~\ref{fig:dataset_construction}.
	To improve reproducibility, we standardized the pipeline from raw collection to final annotation. Raw multimodal pages were processed with structured extraction templates to isolate article text together with linked image or video content. We then removed HyperText Markup Language (HTML) noise, malformed characters, incomplete entries, and duplicate or near-duplicate records. The remaining samples were annotated in Label Studio by trained multilingual annotators under a unified OE schema. In this study, a core opinion is defined as a distinct evaluative statement that is directly relevant to STI monitoring targets, expresses a stance or forward-looking judgment, and represents a distinct opinion within the sample. Ambiguous cases were resolved through iterative review before export to the VL2/VL2.1-compatible JSON format.
	To provide an auxiliary check on sentiment-label consistency, we examined sentiment labels for Chinese (286 samples), Spanish (209 samples), and English (26 samples). Specifically, we compared the finalized human sentiment annotations with independently generated DeepSeek-V3.2 labels under the same positive/neutral/negative schema and computed Cohen's Kappa as a post-hoc agreement indicator. The unweighted Kappa values were $0.5186$ for Chinese, $0.7899$ for Spanish, and $0.7056$ for English, whereas the linearly weighted Kappa values were $0.8325$, $0.9216$, and $0.8710$, respectively. Most disagreements were concentrated in adjacent sentiment categories rather than severe polarity reversals. Together with the manual review, iterative feedback, and consensus-based correction process described above, these results suggest that the sentiment annotations are sufficiently stable and reliable for downstream OE experiments.
	
	The final dataset contains 2,194 OE samples across the three retained modalities and four target languages shown in Table~\ref{tab:modality_language_distribution}. Specifically, the dataset includes 1,198 text samples, 858 image samples, and 138 video samples, with 608 English, 610 Chinese, 482 Russian, and 494 Spanish samples. This diversity supports robust OE experiments by covering varied opinions that are important for intelligence analysis.
	
\begin{table}[!t]
	\centering
	\caption{DISTRIBUTION BY MODALITY AND LANGUAGE}
	\label{tab:modality_language_distribution}
	\begin{tabular}{lr c lr}
		\toprule
		\textbf{Modality} & \textbf{Samples} & \qquad & \textbf{Language} & \textbf{Samples} \\
		\cmidrule{1-2} \cmidrule{4-5}
		Image & 858  & & English & 608 \\
		Video & 138  & & Chinese & 610 \\
		Text  & 1198 & & Russian & 482 \\
		&      & & Spanish & 494 \\
		\cmidrule{1-2} \cmidrule{4-5}
		\textbf{Total} & \textbf{2194} & & \textbf{Total} & \textbf{2194} \\
		\bottomrule
	\end{tabular}
\end{table}
	
	\subsection{Base Model Selection}
	We compared four 7B multimodal LLMs, Video-LLaVA, Video-LLaMA, VideoLLaMA2, and VideoLLaMA2.1, under the practical requirements of our task. The main criteria were support for text, image, and video inputs, available context length, multilingual behavior in English, Chinese, Spanish, and Russian, general performance on relevant benchmarks, and feasibility of training on our available hardware (two A100-40GB GPUs).
	
	From this comparison, we selected VL2 and VL2.1 as the base models. VL2 (7B) combines a CLIP-ViT-Large-Patch14-336 encoder with a Mistral-7B-Instruct-v0.2 decoder and supports contexts of up to 32K tokens. VL2.1 (7B) uses a SigLIP-So400m-Patch14-384 encoder together with a Qwen2-7B-Instruct decoder and extends the context length to 131K tokens. Prior video segmentation research has likewise shown that multi-frame context memory can improve temporal robustness when scene appearance changes across frames \cite{dang2024beyond}. The selected backbones provide multimodal input support, long-context capacity, multilingual decoding, and training feasibility for the STI setting. The comparison is summarized in Table~\ref{tab:model_comparison}.

	\begin{table}[!t]
		\centering
		\footnotesize
		\setlength{\tabcolsep}{3pt}
		\caption{COMPARISON OF MULTIMODAL LARGE LANGUAGE MODELS}
		\label{tab:model_comparison}
		\renewcommand{\arraystretch}{1.3} 
		\begin{tabularx}{\columnwidth}{@{} l c >{\raggedright\arraybackslash\hsize=1.05\hsize}X >{\raggedright\arraybackslash\hsize=0.95\hsize}X @{}}
			\toprule
			\textbf{Model} & \textbf{\makecell{Context\\Length}} & \textbf{Visual Encoder} & \textbf{Language Decoder} \\
			\midrule
			Video-LLaVA & 2K & CLIP Vision Encoder & Vicuna-7B \\
			\addlinespace
			Video-LLaMA & 4K & \makecell[l]{ViT-G/14 +\\ BLIP-2 Q-Former} & LLaMA2-7B \\
			\addlinespace
			VideoLLaMA2 & 32K & \makecell[l]{CLIP-ViT-Large-\\Patch14-336} & \makecell[l]{Mistral-7B-\\Instruct-v0.2} \\
			\addlinespace
			VideoLLaMA2.1 & 131K & \makecell[l]{SigLIP-So400m-\\Patch14-384} & \makecell[l]{Qwen2-7B-\\Instruct} \\
			\bottomrule
		\end{tabularx}
	\end{table}
	
	Taken together, VL2 and VL2.1 offer the most practical balance between context capacity, decoder quality, and training feasibility for the multilingual multimodal OE setting studied here.
	
\subsection{Parameter-Efficient Fine-Tuning Strategies}
To adapt the model efficiently without the memory cost of full-parameter fine-tuning, we adopt QLoRA as the PEFT scheme. QLoRA combines LoRA-style low-rank updates with 4-bit quantization of the frozen backbone \cite{hu2022lora,dettmers2023qlora}, which makes task adaptation practical under limited memory and compute budgets. In our setting, 4-bit NormalFloat (NF4) quantization makes fine-tuning of the multimodal 7B backbones feasible on local hardware such as two A100-40GB GPUs \cite{dettmers2023qlora}.
	
As shown in Fig.~\ref{fig:framework}, QLoRA is attached to representative dense transformations in the transformer blocks. In the schematic, $W_d$ denotes a generic linear transformation to which the low-rank update is applied, including projections in self-attention and feed-forward sublayers. Following the standard LoRA/QLoRA parameterization \cite{hu2022lora,dettmers2023qlora}, the frozen base weight is quantized with NF4 and written as $\widetilde{W}_d$, where $\mathrm{Quant}_{NF4}(\cdot)$ is the NF4 quantization operator. The trainable low-rank factors $A$ and $B$ remain in 16-bit precision, specifically bfloat16 (BF16) or half-precision floating point (FP16). To preserve the behavior of the pretrained model at initialization, we use the standard zero-update setting:

	\begin{equation}
		\widetilde{W}_d = \mathrm{Quant}_{NF4}(W_d)
	\end{equation}
	
	\begin{equation}
		A_0 \sim \mathcal{N}(0,\sigma^2)
	\end{equation}
	
	\begin{equation}
		B_0 = 0
	\end{equation}

	For a representative layer $W_d \in \mathbb{R}^{d \times k}$, the trainable parameters are constrained to low-rank matrices $A$ and $B$, and the weight update $\Delta W_d$ is defined as:
	
	\begin{equation}
		\Delta W_d = B A
	\end{equation}
	
	where $B \in \mathbb{R}^{d \times r}$ and $A \in \mathbb{R}^{r \times k}$ are the low-rank trainable matrices, with rank $r \ll \min(d, k)$, and $d$ and $k$ denote the output and input hidden dimensions of the adapted linear layer, respectively.
	
	During the forward pass, for an input vector $x \in \mathbb{R}^{k}$, the output activation $h \in \mathbb{R}^{d}$ is computed by summing the frozen quantized path and the scaled low-rank adapter path:
	
	\begin{equation}
		h = \widetilde{W}_d x + \frac{\alpha}{r} \Delta W_d x = \widetilde{W}_d x + \frac{\alpha}{r} B A x
	\end{equation}
	
Here, $\alpha$ controls the contribution of the adapter path, and $\sigma^2$ denotes the variance of the Gaussian initialization used for $A_0$. Setting $B_0=0$ ensures that the adapter contributes no update at the start of training, whereas the Gaussian initialization of $A$ provides a stable starting direction for subsequent adaptation \cite{hu2022lora,dettmers2023qlora}. The specific values of the rank $r$ and scaling factor $\alpha$ are determined empirically in the hyperparameter analysis reported later in the experimental section.
	
Training minimizes the standard autoregressive cross-entropy objective over the target token sequence representing the gold JSON core opinions, while updating only the low-rank parameters $\Theta=\{A,B\}$. We optimize these parameters with AdamW. The remaining settings, such as maximum token length, batch size, and epoch count, are reported in the experimental setup. In practice, this configuration reduces memory use enough to make multimodal 7B fine-tuning feasible under our hardware constraints while preserving the standard QLoRA formulation.
\subsection{Case-Level Value Assessment Based on Fuzzy Cumulative Prospect Theory}
Following core-opinion extraction, each predicted opinion is organized with its source, temporal information, and subject to form a case record. We use sentence embeddings to support contextual matching between the predicted opinion and its source/time/subject cues \cite{reimers2019sentence}. We develop a Fuzzy-CPT-based value assessment framework to characterize the intelligence value of each case for STI monitoring, consistent with recent indicator-based quality assessment work for generative AI \cite{yi2026gaihiq}. The framework produces a multidimensional value score and a corresponding case-level category.
\subsubsection{Evaluation Dimensions}
The assessment considers five dimensions: credibility, relevance, local heat, consistency, and recency. Credibility evaluates the authority and evidential characteristics of the source context. Relevance evaluates the semantic association between the extracted opinion and STI priority topics. Local heat captures the strategic attention associated with the topic. Consistency measures the alignment between the opinion and its textual context. Recency reflects the temporal characteristics of the case.
	
\subsubsection{Fuzzy-CPT Fusion}
Fuzzy-CPT integrates the five dimensions into a unified value score \(\mathcal{V}\in[0,1]\), incorporating uncertainty and asymmetric decision preferences \cite{tversky1992prospect,sadeghzadeh2026safe}. Each case is subsequently assigned to one of four value categories: 5-Star Priority, 4-Star High Value, 3-Star Reference, or 2-Star Limited Value. The category thresholds are summarized in Table~\ref{tab:rating_rules}.
	
	\begin{table}[!t]
		\centering
\caption{Fuzzy-CPT Value Categories for Case-Level Assessment}
		\label{tab:rating_rules}
		\begin{tabular}{c c >{\raggedright\arraybackslash}p{0.40\columnwidth}}
			\toprule
			\textbf{Score Range} & \textbf{Stars} & \textbf{Typical Guidance} \\
			\midrule
				\(\mathcal{V} \ge 0.74\) & $\bigstar\bigstar\bigstar\bigstar\bigstar$ & Priority: focused attention. \\
				\(0.61 \le \mathcal{V} < 0.74\) & $\bigstar\bigstar\bigstar\bigstar$ & High value: continued monitoring. \\
				\(0.46 \le \mathcal{V} < 0.61\) & $\bigstar\bigstar\bigstar$ & Reference: supporting context. \\
				\(\mathcal{V} < 0.46\) & $\bigstar\bigstar$ & Limited value: background information. \\
			\bottomrule
		\end{tabular}
	\end{table}
	
\subsubsection{Value Assessment Outputs}
The value assessment results provide a structured representation of each case, including the dimension scores, aggregated value score, star category, and assessment guidance. The resulting JSON record combines the predicted core opinion with contextual information and the corresponding value assessment.

Case-level assessment results can be aggregated by topic, source, and time to support issue watchlists, cross-source analysis, and screening summaries \cite{2}.
\subsection{Evaluation Metrics}
The core-opinion extraction stage is evaluated with four complementary metrics: Precision, Recall, F1-score, and sample-level accuracy. Opinion-level evaluation follows similarity-threshold matching rather than exact string identity.

A predicted core opinion is treated as correctly matched only when it can be paired one-to-one with an unmatched ground-truth core opinion whose textual similarity is no lower than the fixed threshold \(\tau = 0.5\). This threshold provides tolerance for paraphrase and surface-form variation while still preventing loose many-to-one matching. We also report sample-level accuracy as a supplementary indicator of whether a sample contains at least one successfully matched core opinion.

\noindent \textbf{Opinion-Level Metrics} \\
For each sample, let \(P\) denote the set of core-opinion strings predicted by the model and \(G\) denote the set of ground-truth core-opinion strings. Under one-to-one matching with the threshold \(\tau = 0.5\), matched core-opinion pairs are counted as True Positives (TP), unmatched predicted core opinions are counted as False Positives (FP), and unmatched ground-truth core opinions are counted as False Negatives (FN).
Precision, Recall, and F1-score are then calculated as:
\begin{equation}
	\text{Precision} = \frac{TP}{TP + FP}
\end{equation}
\begin{equation}
	\text{Recall} = \frac{TP}{TP + FN}
\end{equation}
\begin{equation}
\text{F1} = \frac{2 \times \text{Precision} \times \text{Recall}}{\text{Precision} + \text{Recall}}.
\end{equation}

Recall measures how many ground-truth core opinions are successfully recovered, whereas Precision reflects how reliable the extracted core opinions are. F1-score summarizes the balance between the two through their harmonic mean.

\noindent \textbf{Sample-Level Accuracy} \\
In addition to opinion-level metrics, we compute sample-level accuracy to reflect coarse sample-level utility. A sample is counted as correct when at least one predicted core opinion is successfully matched to a ground-truth core opinion under the same threshold, namely when \(TP > 0\). Sample-level accuracy is defined as:
\begin{equation}
	\text{SampleAcc} = \frac{\text{Number of samples with } TP > 0}{\text{Total number of samples}}.
	\label{eq:sample_acc}
\end{equation}
This metric indicates whether the model can recover at least one valid core opinion from an input, which is useful for downstream STI analysis that begins with sample-level screening.
	
	\section{Experimental Results and Analysis}
	\subsection{Experimental Setup}
	\subsubsection{Datasets and Evaluation Metrics}
	Experiments are conducted on our multimodal STI dataset, which contains 2,194 samples across four languages (Chinese, English, Spanish, and Russian) and three modalities (text, image, and video). For evaluation, we use a held-out test set of 200 samples, evenly distributed across the four languages, with 50 samples per language. We evaluate core-opinion extraction with four complementary metrics: Precision, Recall, F1-score, and sample-level accuracy. In the STI setting, a successful extraction identifies valid core opinions and returns them in the prescribed JSON format.
	
\subsubsection{Implementation Details}
We adopt VL2 and VL2.1 as the backbones and fine-tune them with QLoRA. Following the hyperparameter analysis reported later, the final configuration uses LoRA rank $r=256$ and scaling factor $\alpha=512$. Unless otherwise specified, the reported benchmark results follow the Image-Augmented setting, in which samples with available static visual evidence are paired with one representative visual cue, whereas text-only samples remain text inputs. The modality ablation reports Text-Centered, Image-Augmented, and Full Modality settings. Text-Centered uses the same multimodal backbone with text-only input, and Full Modality includes text, static images, and temporal videos. The final QLoRA runs are trained for 5 epochs with a maximum token length of 6,000 and a learning rate of $1\times10^{-5}$; the per-device batch size is 2 with gradient accumulation over 4 steps.

	\subsubsection{Prompt-Based Baselines}
	To compare the proposed method against prompt-only adaptation, we evaluate four prompt baselines. Prompt-based reasoning variants have been explored in multimodal reasoning and question-answering settings \cite{12,14,15}. Similar ideas have also been used in knowledge-graph construction and structured long-document generation \cite{22,27}. Zero-Shot prompting ($\sim$465 tokens) uses direct instructions without exemplars and serves as the efficiency baseline. 4-Shot prompting ($\sim$2,991 tokens) adds four cross-lingual examples, one each in English, Chinese, Spanish, and Russian, which improves multilingual consistency and format adherence at a much higher token cost. Chain-of-Thought (CoT, $\sim$622 tokens) asks the model to reason step by step, which can help with subjective identification and more complex extraction decisions \cite{3,4,20,25}. Tree of Thoughts (ToT, $\sim$775 tokens) explores multiple reasoning paths and scores them from 1 to 10, which can be useful in ambiguous cases with implicit opinions \cite{24,25}. Recent work has also examined multi-dimensional evaluation of generated reasoning traces, which further illustrates the diversity of reasoning-oriented outputs \cite{28}. These settings serve as comparative baselines.
	
	\subsection{Zero-Shot Baseline}
	To establish a reference baseline, we evaluate the off-the-shelf VL2 and VL2.1 models without any fine-tuning or specialized prompting. Table~\ref{tab:baseline_performance} reports the zero-shot results for structured core-opinion extraction.

	\begin{table*}[!t]
		\centering
		\caption{ZERO-SHOT BASELINE PERFORMANCE OF VL2 AND VL2.1 ACROSS LANGUAGES}
		\label{tab:baseline_performance}
		\renewcommand{\arraystretch}{1.2}
		\begin{tabular}{llcccc}
			\toprule
			\textbf{Model} & \textbf{Language} & \textbf{Precision (\%)} & \textbf{Recall (\%)} & \textbf{F1-score (\%)} & \textbf{Sample-level Acc (\%)} \\
			\midrule
			\multirow{5}{*}{VL2 (Mistral-based)} 
			& English & 28.70 & 51.20 & 36.78 & 66.00 \\
			& Chinese & 12.11 & 32.29 & 17.61 & 36.00 \\
			& Spanish & 10.58 & 24.18 & 14.72 & 32.00 \\
			& Russian & 4.60 & 9.57 & 6.21 & 10.00 \\
			\cmidrule{2-6}
			& \textbf{Overall} & \textbf{13.82} & \textbf{29.98} & \textbf{18.92} & \textbf{36.00} \\
			\midrule
			\multirow{5}{*}{VL2.1 (Qwen-based)} 
			& English & 24.15 & 68.00 & 35.64 & 84.00 \\
			& Chinese & 17.21 & 65.63 & 27.27 & 66.00 \\
			& Spanish & 3.19 & 9.89 & 4.83 & 16.00 \\
			& Russian & 0.30 & 0.87 & 0.45 & 2.00 \\
			\cmidrule{2-6}
			& \textbf{Overall} & \textbf{11.85} & \textbf{37.00} & \textbf{17.95} & \textbf{42.00} \\
			\bottomrule
		\end{tabular}
	\end{table*}
	\begin{figure}[!t]
		\centering
		\includegraphics[width=\columnwidth]{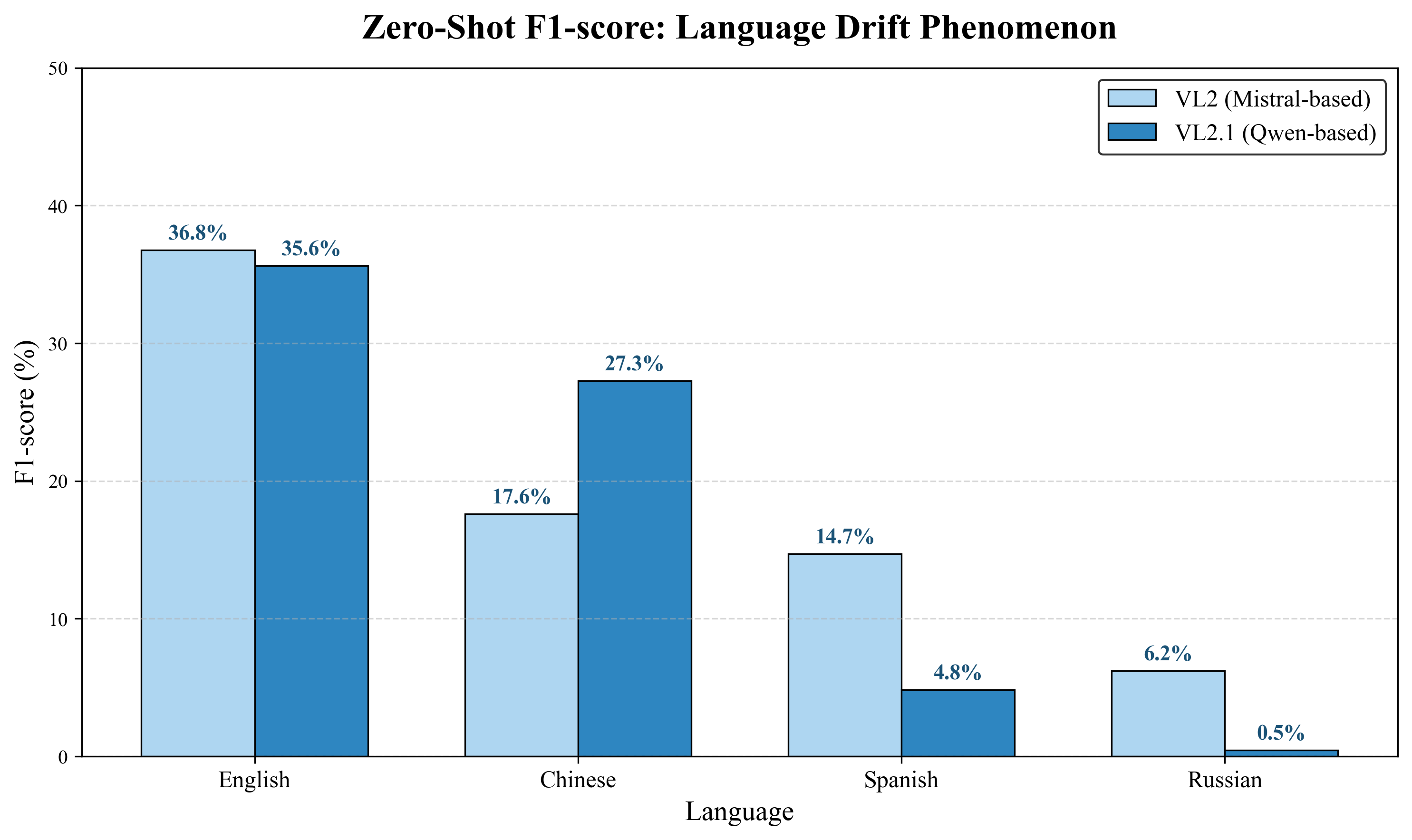}
		\caption{Zero-shot F1-score across four languages.}
		\label{fig:zero_shot_baseline}
	\end{figure}
	\paragraph{Cross-Lingual Zero-Shot Results} As shown in Table~\ref{tab:baseline_performance} and Fig.~\ref{fig:zero_shot_baseline}, the zero-shot setting achieves stronger results in English and Chinese than in Spanish and Russian. For VL2, the F1-score decreases from 36.78\% in English to 14.72\% in Spanish and 6.21\% in Russian. VL2.1 shows the same pattern, reaching 35.64\% F1-score in English, 4.83\% in Spanish, and 0.45\% in Russian. The overall F1-scores of VL2 and VL2.1 are 18.92\% and 17.95\%, respectively.
	
	\paragraph{Precision--Recall Profile of Zero-Shot Extraction} Zero-shot models preserve output structure and retrieve many candidate opinions. For example, VL2.1 reaches 68.00\% Recall and 84.00\% sample-level accuracy in English, with 24.15\% Precision; in Chinese, it reaches 65.63\% Recall, 66.00\% sample-level accuracy, and 17.21\% Precision. The resulting overall F1-scores are 18.92\% for VL2 and 17.95\% for VL2.1. These zero-shot results provide the reference baseline for subsequent adaptation.
	
	\subsection{Prompt-Based Baseline Comparison}
	Before introducing parameter updates, we evaluated whether advanced prompt-based baselines could overcome the extraction defocus and structural inconsistencies observed in the zero-shot setting. We subjected both VL2 and VL2.1 to 4-Shot prompting, Chain-of-Thought (CoT), and Tree of Thoughts (ToT) strategies. The comprehensive results are presented in Table~\ref{tab:prompt_performance}.
	
\begin{figure}[!htbp]
		\centering
		\includegraphics[width=\columnwidth]{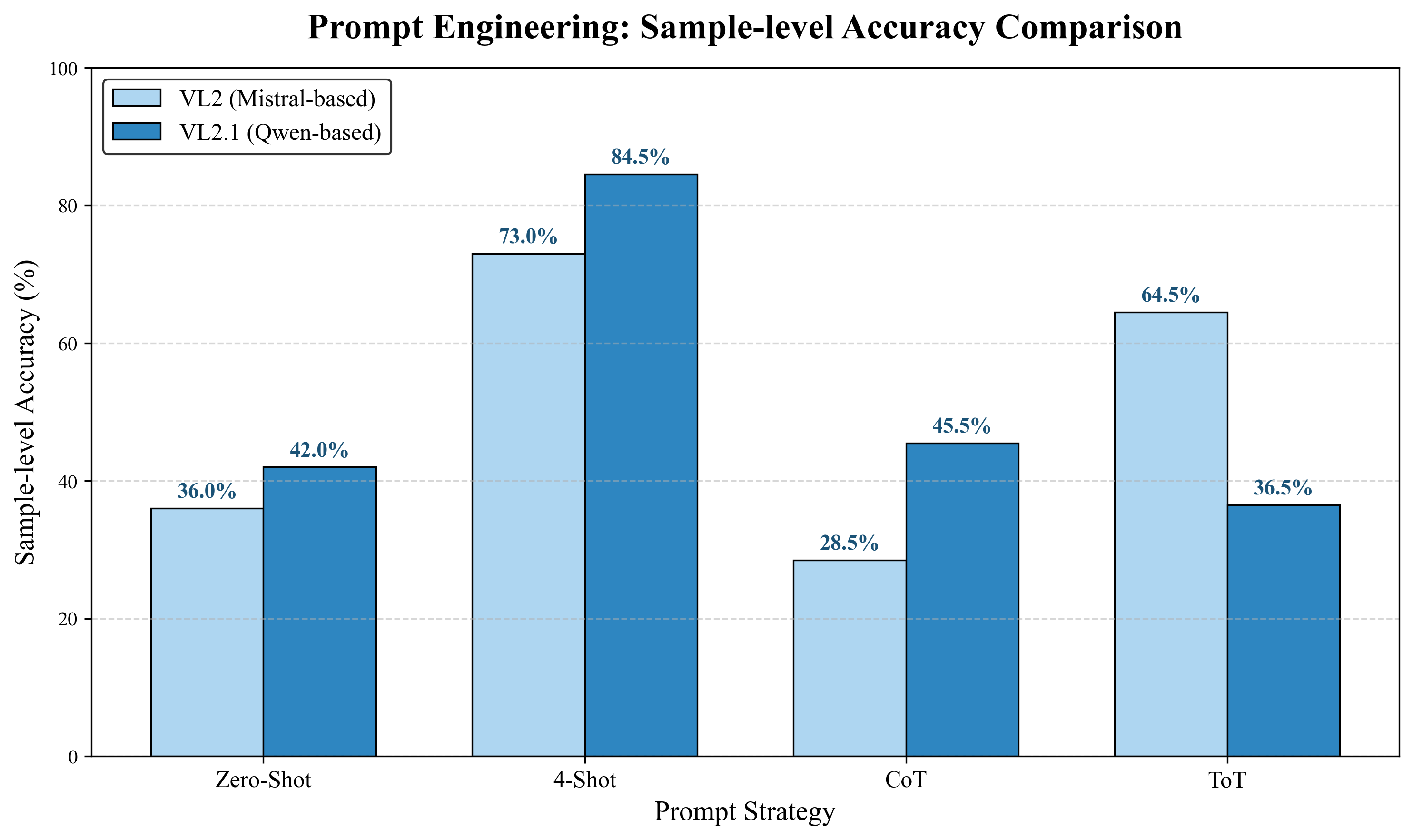}
		\caption{Overall sample-level accuracy comparison across prompt-based baseline strategies.}
		\label{fig:prompt_engineering}
	\end{figure}

	\FloatBarrier
	\begin{table*}[!t]
		\centering
		\caption{PERFORMANCE COMPARISON OF PROMPT ENGINEERING TECHNIQUES ACROSS LANGUAGES}
		\label{tab:prompt_performance}
		\renewcommand{\arraystretch}{1.1}
		\begin{tabular}{lllcccc}
			\toprule
			\textbf{Model} & \textbf{Prompt Strategy} & \textbf{Language} & \textbf{Precision (\%)} & \textbf{Recall (\%)} & \textbf{F1-score (\%)} & \textbf{Sample-level Acc (\%)} \\
			\midrule
			\multirow{20}{*}{VL2} 
			& \multirow{5}{*}{Zero-Shot} 
			& English & 28.70 & 51.20 & 36.78 & 66.00 \\
			& & Chinese & 12.11 & 32.29 & 17.61 & 36.00 \\
			& & Spanish & 10.58 & 24.18 & 14.72 & 32.00 \\
			& & Russian & 4.60 & 9.57 & 6.21 & 10.00 \\
			& & \textbf{Overall} & \textbf{13.82} & \textbf{29.98} & \textbf{18.92} & \textbf{36.00} \\
			\cmidrule{2-7}
			& \multirow{5}{*}{4-Shot} 
			& English & 21.53 & 58.40 & 31.47 & 76.00 \\
			& & Chinese & 24.48 & 61.46 & 35.01 & 72.00 \\
			& & Spanish & 21.45 & 68.13 & 32.63 & 76.00 \\
			& & Russian & 25.19 & 57.39 & 35.01 & 68.00 \\
			& & \textbf{Overall} & \textbf{22.99} & \textbf{60.89} & \textbf{33.38} & \textbf{73.00} \\
			\cmidrule{2-7}
			& \multirow{5}{*}{CoT} 
			& English & 25.93 & 44.80 & 32.84 & 62.00 \\
			& & Chinese & 13.08 & 32.29 & 18.62 & 42.00 \\
			& & Spanish & 2.44 & 4.40 & 3.14 & 6.00 \\
			& & Russian & 0.97 & 1.74 & 1.25 & 4.00 \\
			& & \textbf{Overall} & \textbf{11.30} & \textbf{21.78} & \textbf{14.88} & \textbf{28.50} \\
			\cmidrule{2-7}
			& \multirow{5}{*}{ToT} 
			& English & 28.63 & 56.80 & 38.07 & 74.00 \\
			& & Chinese & 24.53 & 54.17 & 33.77 & 60.00 \\
			& & Spanish & 28.33 & 56.04 & 37.64 & 62.00 \\
			& & Russian & 28.50 & 53.04 & 37.08 & 62.00 \\
			& & \textbf{Overall} & \textbf{27.52} & \textbf{55.04} & \textbf{36.69} & \textbf{64.50} \\
			\midrule
			\multirow{20}{*}{VL2.1} 
			& \multirow{5}{*}{Zero-Shot} 
			& English & 24.15 & 68.00 & 35.64 & 84.00 \\
			& & Chinese & 17.21 & 65.63 & 27.27 & 66.00 \\
			& & Spanish & 3.19 & 9.89 & 4.83 & 16.00 \\
			& & Russian & 0.30 & 0.87 & 0.45 & 2.00 \\
			& & \textbf{Overall} & \textbf{11.85} & \textbf{37.00} & \textbf{17.95} & \textbf{42.00} \\
			\cmidrule{2-7}
			& \multirow{5}{*}{4-Shot} 
			& English & 26.08 & 82.40 & 39.62 & 86.00 \\
			& & Chinese & 20.38 & 79.17 & 32.41 & 82.00 \\
			& & Spanish & 20.47 & 76.92 & 32.33 & 88.00 \\
			& & Russian & 19.44 & 66.09 & 30.04 & 82.00 \\
			& & \textbf{Overall} & \textbf{21.65} & \textbf{76.11} & \textbf{33.71} & \textbf{84.50} \\
			\cmidrule{2-7}
			& \multirow{5}{*}{CoT} 
			& English & 23.53 & 70.40 & 35.27 & 82.00 \\
			& & Chinese & 19.61 & 73.96 & 31.00 & 74.00 \\
			& & Spanish & 3.89 & 16.48 & 6.29 & 22.00 \\
			& & Russian & 0.74 & 2.61 & 1.15 & 4.00 \\
			& & \textbf{Overall} & \textbf{11.57} & \textbf{41.45} & \textbf{18.09} & \textbf{45.50} \\
			\cmidrule{2-7}
			& \multirow{5}{*}{ToT} 
			& English & 26.04 & 40.00 & 31.55 & 48.00 \\
			& & Chinese & 21.56 & 37.50 & 27.38 & 36.00 \\
			& & Spanish & 12.75 & 28.57 & 17.63 & 28.00 \\
			& & Russian & 13.27 & 22.61 & 16.72 & 34.00 \\
			& & \textbf{Overall} & \textbf{18.18} & \textbf{32.32} & \textbf{23.27} & \textbf{36.50} \\
			\bottomrule
		\end{tabular}
	\end{table*}
\paragraph{Few-Shot Prompting Improves Recall and Sample-Level Coverage} For both backbones, 4-Shot is the highest-performing prompt-only setting for Recall and sample-level accuracy; it also gives the highest prompt-only F1-score for VL2.1. Relative to the corresponding zero-shot setting in Table~\ref{tab:prompt_performance}, the 4-Shot F1-score increases from 18.92\% to 33.38\% for VL2 and from 17.95\% to 33.71\% for VL2.1, while sample-level accuracy rises from 36.00\% to 73.00\% for VL2 and from 42.00\% to 84.50\% for VL2.1. The corresponding sample-level trend is also shown in Fig.~\ref{fig:prompt_engineering}. However, Precision remains limited at 22.99\% for VL2 and 21.65\% for VL2.1. These results indicate that few-shot demonstrations increase Recall and sample-level coverage, but are less effective at suppressing irrelevant opinion spans and isolating the core opinion.
	
\paragraph{Prompt-Based Results Across Model Variants} Reasoning-oriented prompts show different results from 4-Shot across the two backbones. On VL2, ToT reaches the best prompt-only F1-score at 36.69\%, while sample-level accuracy is 64.50\%. The VL2 CoT setting reaches 14.88\% F1-score and 28.50\% sample-level accuracy, with Spanish and Russian F1-scores of 3.14\% and 1.25\%, respectively. On VL2.1, CoT reaches 18.09\% F1-score and ToT reaches 23.27\% F1-score with 36.50\% sample-level accuracy. These results show the prompt-based performance profiles across multilingual settings.
\subsection{Impact of QLoRA Hyperparameters}
We next perform multimodal QLoRA fine-tuning. A key step in this stage is selecting the rank ($r$) and scaling factor ($\alpha$) of the low-rank matrices. The rank $r$ determines the capacity of the trainable parameters and therefore affects how much cross-lingual and multimodal task structure the model can absorb.

To isolate the effect of adapter capacity from visual grounding, the hyperparameter search in this section was conducted under the Text-Centered setting on VL2.
	
To identify the optimal configuration, we conducted a comparative ablation study using the VL2 model across three hyperparameter settings: $r=64$ ($\alpha=128$), $r=128$ ($\alpha=256$), and $r=256$ ($\alpha=512$). In all three settings, $\alpha=2r$, so the scaling ratio $\alpha/r=2$ is held constant while adapter capacity changes. The results are detailed in Table~\ref{tab:hyperparameter_tuning}.
	
	\begin{table*}[!t]
		\centering
		\caption{PERFORMANCE COMPARISON OF QLORA HYPERPARAMETER SETTINGS ON VL2 MODEL}
		\label{tab:hyperparameter_tuning}
		\renewcommand{\arraystretch}{1.2}
		\begin{tabular}{lcccccc}
			\toprule
			\textbf{Hyperparameters} & \textbf{Language} & \textbf{Precision (\%)} & \textbf{Recall (\%)} & \textbf{F1-score (\%)} & \textbf{Sample-level Acc (\%)} \\
			\midrule
			\multirow{5}{*}{\makecell{$r=64$ \\ $\alpha=128$}} 
			& English & 40.74 & 26.40 & 32.04 & 56.00 \\
			& Chinese & 68.63 & 36.46 & 47.62 & 68.00 \\
			& Spanish & 44.83 & 28.57 & 34.90 & 52.00 \\
			& Russian & 56.90 & 28.70 & 38.15 & 64.00 \\
			\cmidrule{2-6}
			& \textbf{Overall} & \textbf{51.21} & \textbf{29.74} & \textbf{37.63} & \textbf{60.00} \\
			\midrule
			\multirow{5}{*}{\makecell{$r=128$ \\ $\alpha=256$}} 
			& English & 52.22 & 37.60 & 43.72 & 76.00 \\
			& Chinese & 71.70 & 39.58 & 51.01 & 74.00 \\
			& Spanish & 37.93 & 36.26 & 37.08 & 60.00 \\
			& Russian & 61.29 & 33.04 & 42.94 & 66.00 \\
			\cmidrule{2-6}
			& \textbf{Overall} & \textbf{53.42} & \textbf{36.53} & \textbf{43.39} & \textbf{69.00} \\
			\midrule
			\multirow{5}{*}{\makecell{\textbf{$r=256$} \\ \textbf{$\alpha=512$}}} 
			& English & 51.00 & 40.80 & 45.33 & 78.00 \\
			& Chinese & 54.67 & 42.71 & 47.95 & 70.00 \\
			& Spanish & 49.30 & 38.46 & 43.21 & 66.00 \\
			& Russian & 59.09 & 45.22 & 51.23 & 74.00 \\
			\cmidrule{2-6}
			& \textbf{Overall} & \textbf{53.59} & \textbf{41.92} & \textbf{47.04} & \textbf{72.00} \\
			\bottomrule
		\end{tabular}
	\end{table*}
	
	\paragraph{Higher Rank Mainly Improves Recall} As the LoRA rank increases from $r=64$ to $r=128$ and then to $r=256$, overall Precision changes only modestly from 51.21\% to 53.42\% and 53.59\%, whereas Recall rises steadily from 29.74\% to 36.53\% and 41.92\%. The same monotonic trend appears in overall F1-score, which increases from 37.63\% to 43.39\% and 47.04\%, and in sample-level accuracy, which improves from 60.00\% to 69.00\% and 72.00\%. This pattern indicates that increasing adapter capacity improves coverage of valid core opinions.
	
\paragraph{Language-Level Results of the Selected Configuration} The setting with $r=256$ and $\alpha=512$ yields the best overall result among the tested configurations. The F1-score of Russian improves from 38.15\% at $r=64$ to 51.23\% at $r=256$, and the F1-score of English rises from 32.04\% to 45.33\%. Spanish also reaches its best F1-score at $r=256$, with 43.21\%. Chinese peaks at 51.01\% F1-score under $r=128$ and reaches 47.95\% at $r=256$, where Recall rises and Precision decreases from 71.70\% to 54.67\%. The selected configuration achieves the highest overall F1-score among the evaluated settings.
	
\subsection{Performance of Multimodal QLoRA Fine-Tuning}
With the hyperparameters fixed at $r=256$ and $\alpha=512$, we fine-tune both VL2 and VL2.1 under the selected Image-Augmented setting. The goal of this stage is to improve cross-lingual selectivity in core-opinion extraction beyond what prompt-based baselines can achieve. The final fine-tuning results are presented in Table~\ref{tab:finetuning_performance}.
	
\begin{figure}[!t]
		\centering
		\includegraphics[width=\columnwidth]{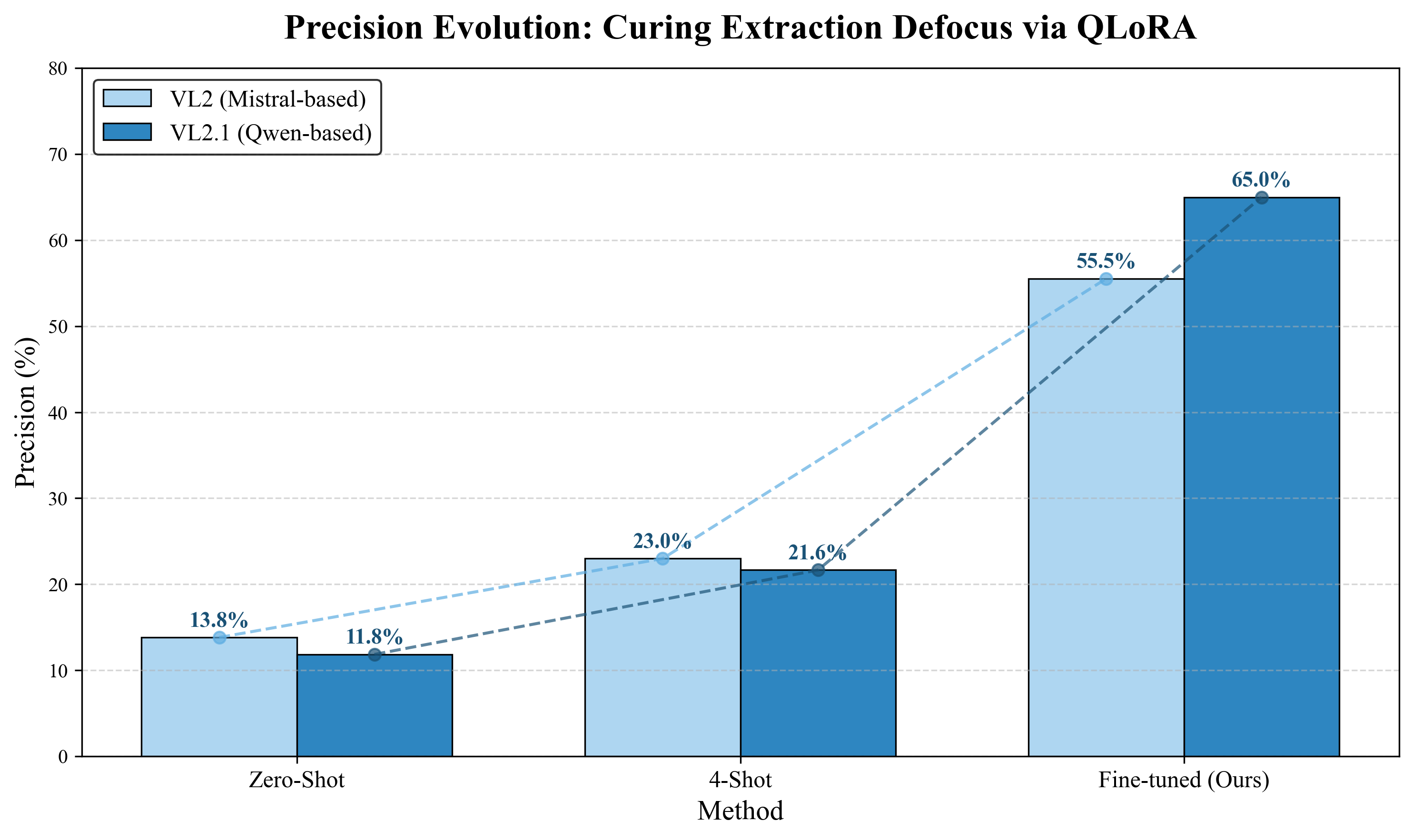}
		\caption{Overall precision comparison among the zero-shot, 4-Shot, and QLoRA fine-tuning settings.}
		\label{fig:qlora_transformation}
	\end{figure}

	\begin{table*}[!t]
		\centering
		\caption{PERFORMANCE OF THE SELECTED IMAGE-AUGMENTED QLORA FINE-TUNING ACROSS LANGUAGES}
		\label{tab:finetuning_performance}
		\renewcommand{\arraystretch}{1.2}
		\begin{tabular}{llcccc}
			\toprule
			\textbf{Model} & \textbf{Language} & \textbf{Precision (\%)} & \textbf{Recall (\%)} & \textbf{F1-score (\%)} & \textbf{Sample-level Acc (\%)} \\
			\midrule
			\multirow{5}{*}{Fine-tuned VL2} 
			& English & 56.52 & 41.60 & 47.93 & 78.00 \\
			& Chinese & 58.33 & 43.75 & 50.00 & 74.00 \\
			& Spanish & 53.52 & 41.76 & 46.91 & 70.00 \\
			& Russian & 54.00 & 46.96 & 50.23 & 78.00 \\
			\cmidrule{2-6}
			& \textbf{Overall} & \textbf{55.52} & \textbf{43.56} & \textbf{48.82} & \textbf{75.00} \\
			\midrule
			\multirow{5}{*}{Fine-tuned VL2.1} 
			& English & 60.22 & 44.80 & 51.38 & 88.00 \\
			& Chinese & 73.68 & 43.75 & 54.90 & 74.00 \\
			& Spanish & 57.38 & 38.46 & 46.05 & 62.00 \\
			& Russian & 71.21 & 40.87 & 51.93 & 72.00 \\
			\cmidrule{2-6}
			& \textbf{Overall} & \textbf{64.98} & \textbf{42.15} & \textbf{51.14} & \textbf{74.00} \\
			\bottomrule
		\end{tabular}
	\end{table*}
\paragraph{Parameter Updates Improve Precision and F1-Score Relative to Prompt-Only Adaptation} Compared with the highest-F1 prompt-only result of VL2.1 under 4-Shot, QLoRA fine-tuning increases Precision from 21.65\% to 64.98\% and F1-score from 33.71\% to 51.14\%, while Recall decreases from 76.11\% to 42.15\%. Sample-level accuracy also decreases from 84.50\% to 74.00\%. The precision trajectory across the baseline and fine-tuning stages is visualized in Fig.~\ref{fig:qlora_transformation}. The results show a precision-oriented shift after parameter updates, with higher Precision and F1-score and lower Recall and sample-level accuracy.
	
	\paragraph{VL2.1 Provides The Better Overall Precision--Recall Balance} Under the same QLoRA configuration, VL2.1 improves overall Precision from 55.52\% to 64.98\% and F1-score from 48.82\% to 51.14\% relative to VL2, whereas VL2 retains slightly higher Recall and sample-level accuracy at 43.56\% and 75.00\% versus 42.15\% and 74.00\%. The advantage of VL2.1 is especially clear in Chinese, where it reaches 73.68\% Precision and 54.90\% F1-score, compared with 58.33\% Precision and 50.00\% F1-score for VL2. The gap is smaller in Russian, where the two models reach F1-scores of 51.93\% and 50.23\%. These results indicate that the Qwen-based VL2.1 backbone remains more selective overall, even though the Mistral-based VL2 still preserves slightly broader coverage.
	
\paragraph{Residual Error Analysis} We manually inspected the 50 English test cases produced by the selected Image-Augmented VL2.1 model. The model recovers at least one valid core opinion in 44 of the 50 cases, consistent with the 88.00\% sample-level accuracy reported in Table~\ref{tab:finetuning_performance}. Among these English cases, 12 are clean recoveries. Observed errors include partial misses, peripheral additions, and incomplete recovery of dense multi-opinion passages.
	
	\paragraph{Dense Multi-Opinion Inputs Remain Challenging} The inspected English cases reveal three recurring patterns. First, in multi-opinion inputs the model often preserves the central strategic statement while omitting secondary but still valid core opinions. For example, in an Amazon sustainability sample, the model retains the business-strategy commitment statement but misses both the 2040 net-zero target and the contrastive remark about criticism. Second, residual over-extraction remains: in a Starbucks sample, the model captures the backlash theme, but fragments it into short items such as ``environmental groups'' and ``backlash'' while also adding a non-gold statement about reusable alternatives. Third, a small number of cases still drift toward a topically related but incorrect statement; in one Boeing sample, the prediction shifts to a broad sustainability quotation and misses the gold opinions on anti-corruption and stakeholder engagement. These observations indicate that the remaining challenge is no longer coarse output formatting, but selective recovery of complete opinion sets from rhetorically dense passages.
	
	\FloatBarrier
	\subsection{Ablation Study on Modalities}
	To examine how additional visual evidence affects extraction quality under the experimental setup used in this paper, we conducted an auxiliary ablation study on the test set with both fine-tuned VL2 and VL2.1.
	
\begin{figure}[!t]
		\centering
		\includegraphics[width=\columnwidth]{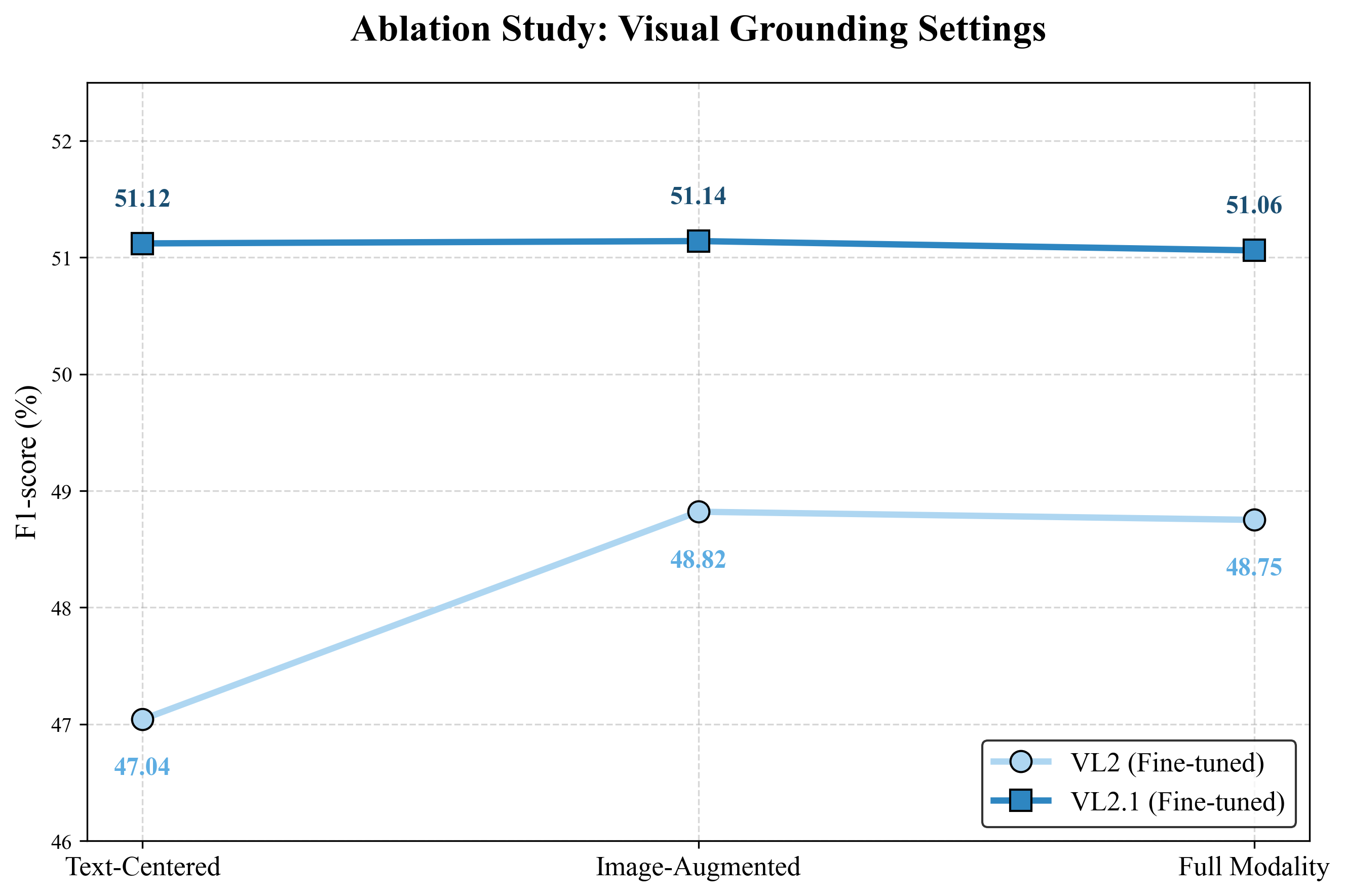}
		\caption{Ablation study comparing Text-Centered, Image-Augmented, and Full Modality visual grounding settings.}
		\label{fig:modality_ablation}
	\end{figure}

	We compared three settings. \textbf{Text-Centered} uses the same multimodal backbone with text-only input and no visual cue. \textbf{Image-Augmented} pairs each sample with one representative static visual cue and is adopted as the main benchmark setting in this paper. \textbf{Full Modality} uses the complete heterogeneous input, including text, static images, and temporal videos, and is retained here as a supplementary analysis. The comparative results are presented in Table~\ref{tab:modality_ablation} and Fig.~\ref{fig:modality_ablation}.
	
	\begin{table}[!t]
		\centering
		\footnotesize
		\setlength{\tabcolsep}{3pt}
		\caption{ABLATION STUDY ON VISUAL GROUNDING SETTINGS ACROSS DIFFERENT MODELS}
		\label{tab:modality_ablation}
		\renewcommand{\arraystretch}{1.15}
		\begin{tabularx}{\columnwidth}{@{}l>{\raggedright\arraybackslash}Xccc@{}}
			\toprule
			\textbf{Model} & \textbf{\makecell[l]{Modality\\Setting}} & \textbf{Prec. (\%)} & \textbf{Rec. (\%)} & \textbf{F1 (\%)} \\
			\midrule
			\multirow{3}{*}{VL2} 
			& Text-Centered & 53.59 & 41.92 & 47.04 \\
			& \textbf{Image-Augmented} & \textbf{55.52} & \textbf{43.56} & \textbf{48.82} \\
			& Full Modality & 55.36 & 43.56 & 48.75 \\
			\midrule
			\multirow{3}{*}{VL2.1} 
			& Text-Centered & 63.32 & 42.86 & 51.12 \\
			& \textbf{Image-Augmented} & \textbf{64.98} & 42.15 & \textbf{51.14} \\
			& Full Modality & 64.75 & 42.15 & 51.06 \\
			\bottomrule
		\end{tabularx}
	\end{table}
\paragraph{Image-Augmented Inputs Are the Selected Main Configuration} Moving from the Text-Centered setting to Image-Augmented inputs raises VL2 from 47.04\% to 48.82\% in F1-score and from 53.59\% to 55.52\% in Precision. VL2.1 shows the same tendency: Precision increases from 63.32\% to 64.98\%, while F1-score is maintained at 51.14\% versus 51.12\%. Image-Augmented inputs are selected as the main benchmark setting based on these results.
	
\paragraph{Full Modality Results} The Full Modality condition achieves 48.75\% F1-score for VL2 and 51.06\% for VL2.1, compared with 48.82\% and 51.14\% in the Image-Augmented setting. This comparison is consistent with prior video modeling work on temporal context management \cite{dang2024adaptive}. The Full Modality results are reported as a supplementary modality comparison.
	
	\FloatBarrier
	\subsection{Comparison with External Multimodal Models}
\begin{figure}[!t]
		\centering
		\includegraphics[width=\columnwidth]{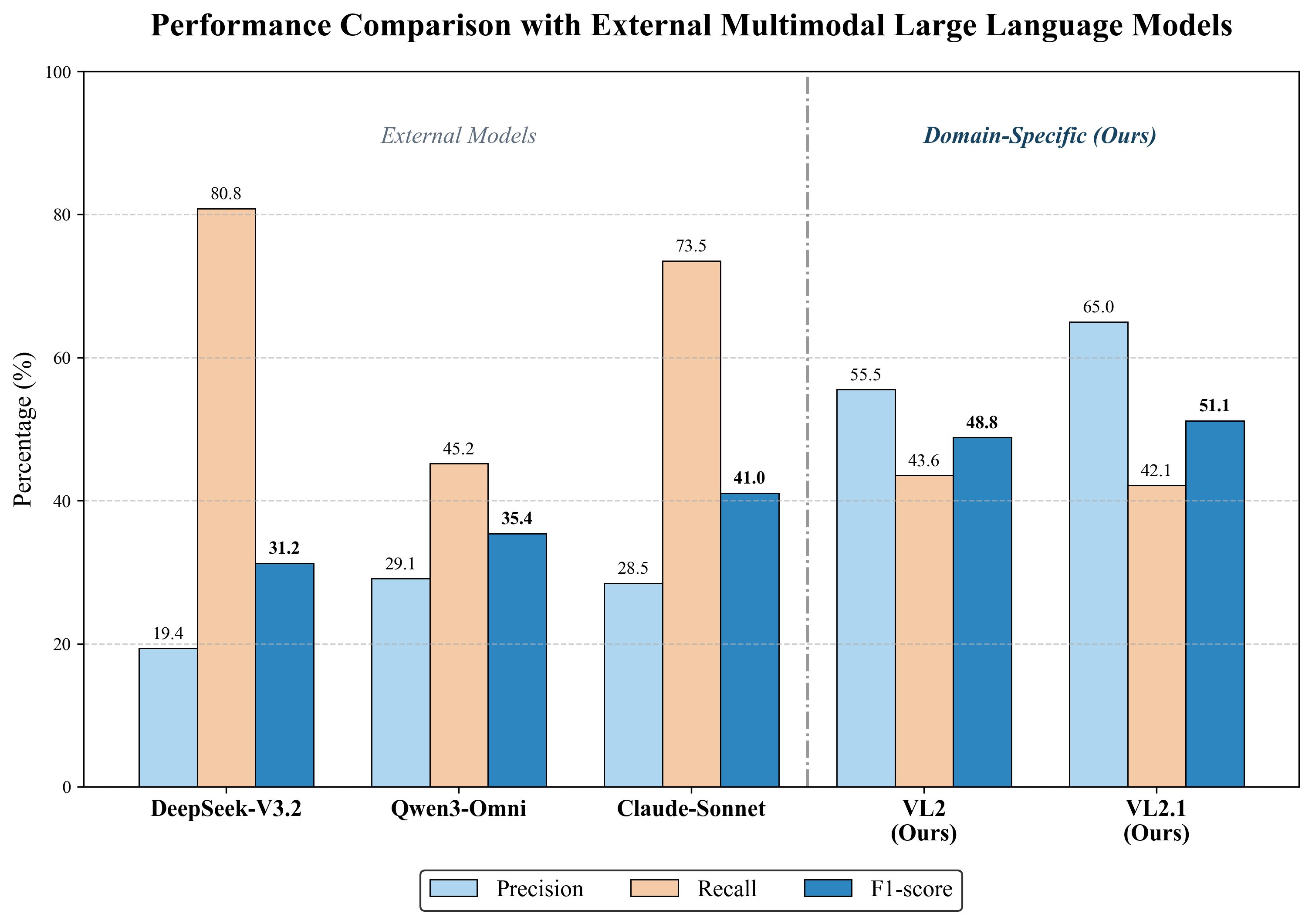}
		\caption{Performance comparison with external multimodal large language models. }
		\label{fig:sota_comparison}
	\end{figure}

	To compare the proposed multimodal QLoRA fine-tuning framework with external models, we evaluate DeepSeek-V3.2, Qwen3-Omni-Flash (VideoFrames), and Claude-Sonnet-4-6. All models are evaluated with the same zero-shot prompts and the same strict structured-output constraints. The overall comparison is reported in Table~\ref{tab:sota_comparison} and Fig.~\ref{fig:sota_comparison}.
	
	\begin{table*}[!t]
		\centering
		\caption{PERFORMANCE COMPARISON WITH EXTERNAL MULTIMODAL LARGE LANGUAGE MODELS}
		\label{tab:sota_comparison}
		\renewcommand{\arraystretch}{1.2}
		\begin{tabular}{lcccc}
			\toprule
			\textbf{Model} & \textbf{Precision (\%)} & \textbf{Recall (\%)} & \textbf{F1-score (\%)} & \textbf{Sample-level Acc (\%)} \\
			\midrule
			DeepSeek-V3.2 & 19.37 & 80.80 & 31.25 & 85.50 \\
			Qwen3-Omni-Flash & 29.11 & 45.20 & 35.41 & 12.50 \\
			Claude-Sonnet-4-6 & 28.46 & 73.53 & 41.04 & 15.50 \\
			\midrule
			\textbf{Ours (Fine-tuned VL2)} & 55.52 & 43.56 & 48.82 & 75.00 \\
			\textbf{Ours (Fine-tuned VL2.1)} & \textbf{64.98} & 42.15 & \textbf{51.14} & 74.00 \\
			\bottomrule
		\end{tabular}
	\end{table*}
	\paragraph{Fine-Tuned VL2.1 Results in External Comparison} Among all compared systems, fine-tuned VL2.1 achieves the highest Precision at 64.98\% and the highest F1-score at 51.14\%. The highest-F1 external baseline is Claude-Sonnet-4-6 at 41.04\%, giving a difference of 10.10 percentage points; relative to DeepSeek-V3.2 at 31.25\%, the difference is 19.89 points. The comparison reports the performance of the external systems and the STI-specific fine-tuned model under the shared evaluation setting.
	
	\paragraph{External Model Results} DeepSeek-V3.2 achieves 19.37\% Precision, 80.80\% Recall, 31.25\% F1-score, and 85.50\% sample-level accuracy. Claude-Sonnet-4-6 achieves 28.46\% Precision, 73.53\% Recall, and 41.04\% F1-score. Qwen3-Omni-Flash achieves 29.11\% Precision, 45.20\% Recall, and 35.41\% F1-score. These results provide a comparative reference for the fine-tuned VL2.1 model.
\subsection{Value Assessment Demonstration}
We apply the proposed value assessment framework to 289 predicted opinion-level cases. Each case combines a predicted core opinion with its source, temporal information, and subject context. The framework evaluates the five dimensions and produces a value score and star category for each case.

\begin{figure}[!htbp]
		\centering
		\includegraphics[width=\columnwidth]{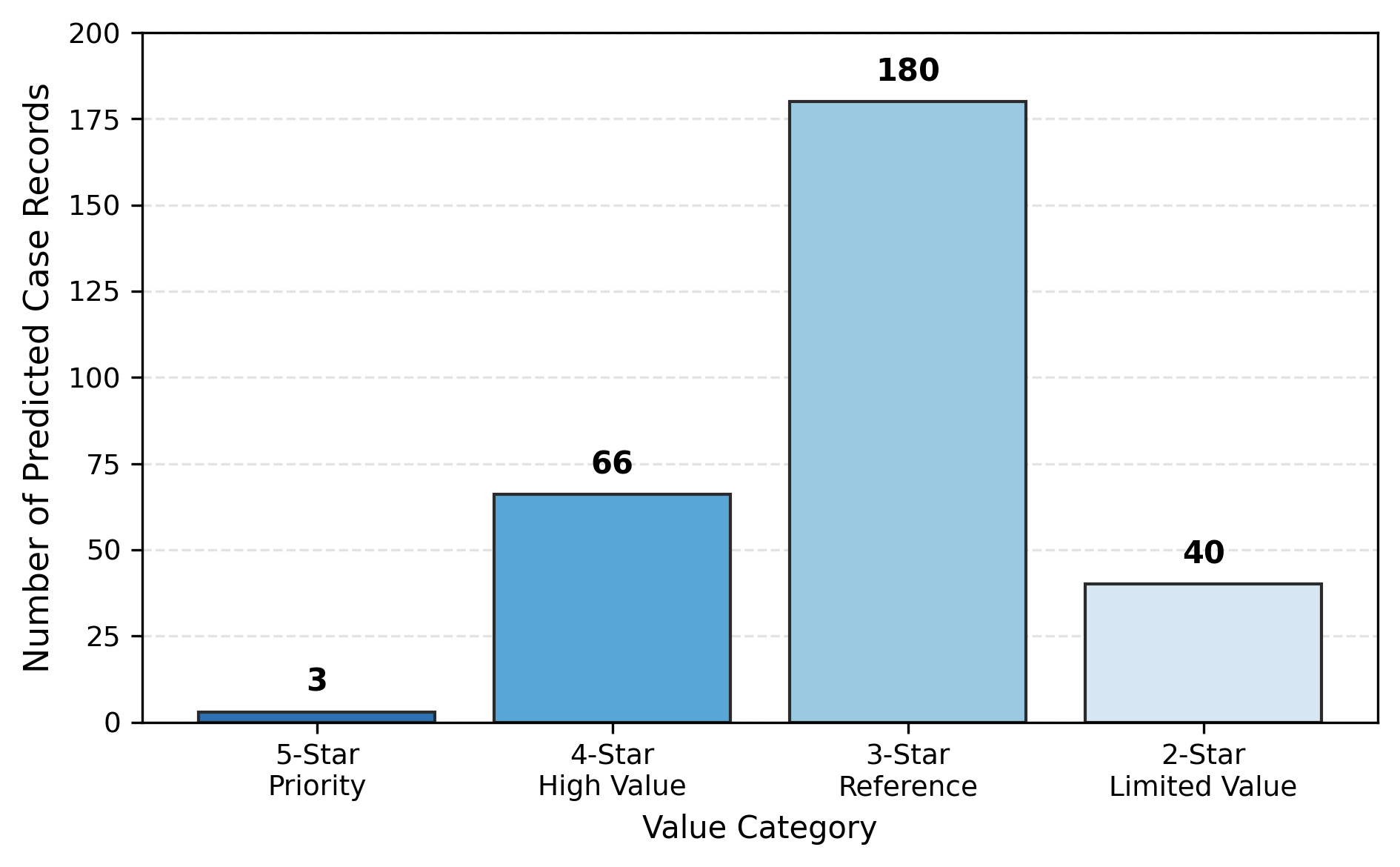}
		\caption{Distribution of value categories among predicted opinion-level case records.}
		\label{fig:value_distribution}
	\end{figure}

\paragraph{Value Category Distribution} As shown in Fig.~\ref{fig:value_distribution}, 180 of the 289 predicted opinion-level cases are categorized as 3-Star Reference, 66 are assigned to 4-Star High Value, 3 reach 5-Star Priority, and 40 fall into the 2-Star Limited Value category. The distribution provides a structured overview of the value characteristics of the extracted opinions.

\paragraph{Topic-Level Value Summary} The 69 cases categorized as 4-Star High Value or 5-Star Priority are summarized by topic and source coverage in Table~\ref{tab:topic_watchlist}. Frontier AI, electric and autonomous vehicles, clean energy transition, and digital transformation emerge as recurrent high-value topics.

\begin{table}[!htbp]
\centering
\caption{Topic-Level Summary of High-Value Cases}
\label{tab:topic_watchlist}
\footnotesize
\begin{tabular}{lcc}
\toprule
\textbf{Topic} & \textbf{4/5-Star Records} & \textbf{Source Types} \\
\midrule
Frontier AI & 17 & 5 \\
\makecell[l]{Electric and Autonomous\\Vehicles} & 9 & 4 \\
\makecell[l]{Clean Energy\\Transition} & 7 & 3 \\
Digital Transformation & 6 & 4 \\
\bottomrule
\end{tabular}
\end{table}

\paragraph{Illustrative Case} For an English test sample on the growth of India's IT sector, the case is characterized by Source = Industry Report, Time = ``fiscal year 2024'', Subject = Digital Transformation, and Opinion = ``We are witnessing a digital revolution in India''. The Fuzzy-CPT module assigns this case a value score of 0.8109 and categorizes it as 5-Star Priority.
	
\paragraph{Value Assessment for STI Monitoring} The value assessment module integrates the extracted opinion with its contextual information to provide a structured representation of case value. The resulting categories support the organization, comparison, and monitoring of STI-related opinions.
	\FloatBarrier
	\section{Conclusion}
	This study presents a technical framework for multimodal and multilingual core-opinion extraction in STI. By integrating multimodal contextual anchoring, QLoRA-based PEFT, and case-level value assessment, the framework is designed to address language drift and the computational cost associated with traditional zero-shot MLLMs.
	
	On the held-out 200-sample test set drawn from the self-constructed STI dataset of 2,194 samples, fine-tuned VideoLLaMA2.1 under the selected Image-Augmented setting achieves 64.98\% Precision, 42.15\% Recall, 51.14\% F1-score, and 74.00\% sample-level accuracy. For Russian and Spanish, the F1-scores increase from 0.45\%--4.83\% in the zero-shot VL2.1 setting to 46.05\%--51.93\% after fine-tuning. The comparison between Text-Centered, Image-Augmented, and Full Modality settings shows that visual grounding provides a modest auxiliary cue for core-opinion identification. Among the tested settings, $r=256$ and $\alpha=512$ are selected on the basis of the best overall result for domain-specific adaptation in this study.
	
The proposed framework offers a parameter-efficient approach for automated global intelligence monitoring. By producing structured core opinions together with contextual case information and value assessment results, it can support integration with existing STI databases and downstream policy analysis in complex cross-domain information environments.
	Future work will extend the present case-level assessment scheme toward topic-level trend synthesis and broader cross-platform STI monitoring.
	
	\section*{Acknowledgment}
	This work is supported by the National Key Research and Development Program [2022YFB3103602]. The authors would also like to express their sincere gratitude to Yuhan Tu and Xuecheng Hou for their valuable assistance and support throughout this research.
	


\begin{thebibliography}{99}



		\bibitem{1} J. Wu, M. Jiang, J. Fan, S. Li, H. Xu, Y. Zhao, ``Arch-eval benchmark for assessing Chinese architectural domain knowledge in large language models,'' \textit{Scientific Reports}, vol. 15, no. 1, pp. 13485--13485, 2025.




		

		\bibitem{5} Z. Chen, F. Imani, ``A multi-expert framework for enhancing multimodal large language models in industrial anomaly detection,'' \textit{Pattern Recognition}, vol. 172, pp. 112752--112752, 2026.




		

		\bibitem{26} W. Wang, H. Gu, Z. Wu, H. Chen, X. Chen, F. Shi, ``Ptfusion: LLM-driven context-aware knowledge fusion for web penetration testing,'' \textit{Information Fusion}, vol. 127, pp. 103731--103731, 2026.




		
		\bibitem{hong2026aigcvideo} S. Hong, X. Wang, C. Zhang, J. Wang, P. Duan, Y. Wang, ``AIGC video detection based on the fusion of spatial-frequency-optical flow multimodal features,'' \textit{Journal of Systems Engineering and Electronics}, pp. 1--15, 2026, doi: 10.23919/JSEE.2026.000049.



		\bibitem{dang2023efficient} J. Dang, H. Zheng, J. Lai, X. Yan, Y. Guo, ``Efficient and robust video object segmentation through isogenous memory sampling and frame relation mining,'' \textit{IEEE Transactions on Image Processing}, vol. 32, pp. 3924--3938, 2023.




		

		\bibitem{8} Z. A. Naci, M. R. Hossain, F. A. Mamun, ``Evaluation of open and closed-source LLMs for low-resource language with zero-shot, few-shot, and chain-of-thought prompting,'' \textit{Natural Language Processing Journal}, vol. 10, pp. 100124--100124, 2025.




		

		\bibitem{9} D. T. Do, M. P. Nguyen, L. M. Nguyen, ``Enhancing zero-shot multilingual semantic parsing: A framework leveraging large language models for data augmentation and advanced prompting techniques,'' \textit{Neurocomputing}, vol. 618, pp. 129108--129108, 2025.




		

		\bibitem{7} H. Chen, J. Wang, W. Wang, Y. Xu, ``Improving zero-shot chain-of-thought reasoning across languages with rectification and self-optimization prompting,'' \textit{The Journal of Supercomputing}, vol. 81, no. 10, pp. 1096--1096, 2025.




		

		\bibitem{10} S. Chen, W. Wang, X. Chen, P. Lu, Z. Yang, Y. Du, ``Llama-lora neural prompt engineering: A deep tuning framework for automatically generating Chinese text logical reasoning thinking chains,'' \textit{Data Intelligence}, vol. 6, no. 2, pp. 375--408, 2024.




		

		\bibitem{16} X. Wang, Z. Xu, Y. Zheng, H. Wang, ``Parameter-efficient weakly supervised referring video object segmentation via chain-of-thought reasoning,'' \textit{Complex \& Intelligent Systems}, vol. 11, no. 6, pp. 273--273, 2025.



		\bibitem{dang2025fwise} J. Dang, H. Zheng, B. Wang, J. Li, H. Ding, J. Lai, ``Efficient video object segmentation based on frame-wise and segment-wise spatio-temporal interaction memory networks,'' \textit{Scientia Sinica Informationis}, vol. 55, no. 1, pp. 80--93, 2025.




		

		\bibitem{18} B. S. Rathore, S. Chaurasia, ``Fine tuning large language models for hate speech detection in high-risk and code mixed custom dataset through a socially responsible approach for safer digital platforms,'' \textit{Discover Sustainability}, vol. 6, no. 1, pp. 1409--1409, 2025.




		

		\bibitem{19} R. Pan, J. A. G. D\'{\i}az, R. V. Garc\'{\i}a, ``Spanish mltlhecorpus 2023: Multi-task learning for hate speech detection to identify speech type, target, target group and intensity,'' \textit{Computer Standards \& Interfaces}, vol. 94, pp. 103990--103990, 2025.




		

		\bibitem{31} Y. Liu, S. Y. M. Lee, D. Li, ``Examining emotions in English and translated Chinese children's literature: a bilingual emotion detection model based on LLMs,'' \textit{Language Resources and Evaluation}, vol. 59, no. 4, pp. 1--33, 2025.



		\bibitem{dang2024beyond} J. Dang, H. Zheng, X. Xu, L. Wang, Y. Guo, ``Beyond appearance: Multi-frame spatio-temporal context memory networks for efficient and robust video object segmentation,'' \textit{IEEE Transactions on Image Processing}, vol. 33, pp. 4853--4866, 2024.




		

		\bibitem{hu2022lora} E. J. Hu, Y. Shen, P. Wallis, Z. Allen-Zhu, Y. Li, S. Wang, L. Wang, W. Chen, ``LoRA: Low-Rank Adaptation of Large Language Models,'' in \textit{Proc. Int. Conf. Learn. Representations (ICLR)}, 2022.




		
		\bibitem{dettmers2023qlora} T. Dettmers, A. Pagnoni, A. Holtzman, L. Zettlemoyer, ``QLoRA: Efficient Finetuning of Quantized LLMs,'' in \textit{Advances in Neural Information Processing Systems}, vol. 36, 2023.




		
		\bibitem{yi2026gaihiq} J. Yi, F. Du, Y. Nie, W. Liang, X. Zhou, J. Chen, G. Li, M. Liu, Y. Lv, W. Zhao, X. Hou, ``GAI-HIQ: Developing a health information quality assessment indicator system for generative artificial intelligence,'' \textit{Information Processing \& Management}, vol. 63, no. 5, pp. 104651--104651, 2026.




		

		\bibitem{sadeghzadeh2026safe} N. Sadeghzadeh, M. Robati, S. M. Monavari, K. Ziari, ``A comprehensive sustainability evaluation through the adaptive SAFE-fuzzy model: a methodology toward urban area,'' \textit{Environment, Development and Sustainability}, pp. 1--39, 2026.





		\bibitem{reimers2019sentence} N. Reimers, I. Gurevych, ``Sentence-BERT: Sentence Embeddings using Siamese BERT-Networks,'' in \textit{Proceedings of the 2019 Conference on Empirical Methods in Natural Language Processing and the 9th International Joint Conference on Natural Language Processing (EMNLP-IJCNLP)}, Hong Kong, China, pp. 3982--3992, 2019.




		

		\bibitem{tversky1992prospect} A. Tversky, D. Kahneman, ``Advances in prospect theory: Cumulative representation of uncertainty,'' \textit{Journal of Risk and Uncertainty}, vol. 5, pp. 297--323, 1992.




		

		\bibitem{2} X. Xiao, Y. Li, X. He, J. Fang, Z. Yan, C. Xie, ``An assessment framework of higher-order thinking skills based on fine-tuned large language models,'' \textit{Expert Systems with Applications}, vol. 272, pp. 126531--126531, 2025.




		

		\bibitem{12} X. Wen, H. Wang, K. Chen, T. Hu, G. Chen, ``Gmcot: a graph-augmented multimodal chain-of-thought reasoning framework for multi-label zero-shot learning,'' \textit{Frontiers of Information Technology \& Electronic Engineering}, vol. 26, no. 12, pp. 2623--2637, 2025.




		

		\bibitem{14} G. Huang, Y. Long, C. Luo, ``Improving multi-hop question answering with prompting explicit and implicit knowledge aligned human reading comprehension,'' \textit{International Journal of Machine Learning and Cybernetics}, vol. 16, no. 10, pp. 1--16, 2025.




		

		\bibitem{15} M. Tang, C. Bian, L. Yang, X. Zhong, ``Key-concept thinking prompting for improved reasoning in large language models,'' \textit{Neurocomputing}, vol. 656, pp. 130986--130986, 2025.




		

		\bibitem{22} X. Qi, B. Yang, S. Wang, Z. Zhang, Y. Zhang, K. Du, ``Few-shot and chain-of-thought prompting for equipment maintenance knowledge graph construction via large language models,'' \textit{Knowledge-Based Systems}, vol. 335, pp. 115266--115266, 2026.




		

		\bibitem{27} X. Chen, Z. Chen, S. Cheng, ``Cothssum: Structured long-document summarization via chain-of-thought reasoning and hierarchical segmentation,'' \textit{Journal of King Saud University Computer and Information Sciences}, vol. 37, no. 4, pp. 40--40, 2025.




		

		\bibitem{3} Y. Feng, W. An, H. Wang, Z. Yin, ``Enhancing scientific literature summarization via contrastive learning and chain-of-thought prompting,'' \textit{Scientometrics}, vol. 130, no. 8, pp. 1--27, 2025.




		

		\bibitem{4} W. Xu, M. S. S. Kassim, W. L. Hoo, W. Yang, T. Xu, ``Explainable AI for education: Enhancing essay scoring via rubric-aligned chain-of-thought prompting,'' \textit{International Journal of Modern Physics C}, vol. 37, no. 06, 2025.




		

		\bibitem{20} D. Ding, X. Fu, X. Peng, X. Fan, H. Huang, B. Zhang, ``Leveraging chain-of-thought to enhance stance detection with prompt-tuning,'' \textit{Mathematics}, vol. 12, no. 4, 2024.




		

		\bibitem{25} K. Hebenstreit, R. Praas, L. P. Kiesewetter, M. Samwald, ``A comparison of chain-of-thought reasoning strategies across datasets and models,'' \textit{PeerJ Computer Science}, vol. 10, pp. 1999--1999, 2024.




		

		\bibitem{24} Q. Pan, W. Ji, Y. Ding, J. Li, S. Chen, J. Wang, J. Zhou, Q. Chen, M. Zhang, Y. Wu, L. He, ``A survey of slow thinking-based reasoning LLMs using reinforcement learning and test-time scaling law,'' \textit{Information Processing and Management}, vol. 63, no. 2PA, pp. 104394--104394, 2026.




		

		\bibitem{28} L. F. B. Monsalve, G. S. Torres, J. W. B. Bedoya, ``Multi-dimensional evaluation of auto-generated chain-of-thought traces in reasoning models,'' \textit{AI}, vol. 7, no. 1, pp. 35--35, 2026.



		\bibitem{dang2024adaptive} J. Dang, H. Zheng, X. Xu, L. Wang, Q. Hu, Y. Guo, ``Adaptive sparse memory networks for efficient and robust video object segmentation,'' \textit{IEEE Transactions on Neural Networks and Learning Systems}, vol. 36, no. 2, pp. 3820--3833, 2025.

\end{thebibliography}
\end{document}